\documentclass[10pt,twocolumn,letterpaper]{article}

\usepackage{iccv}
\usepackage{times}
\usepackage{epsfig}
\usepackage{graphicx}
\usepackage{amsmath}
\usepackage{amssymb}

\usepackage[accsupp]{axessibility}
\usepackage{authblk}
\usepackage{booktabs}
\usepackage[dvipsnames]{xcolor}
\usepackage{colortbl}
\usepackage{color} 
\usepackage[normalem]{ulem}
\usepackage{multirow}
\usepackage{graphicx}
\usepackage{color}

\def\fig#1{Figure \ref{fig:#1}}

\def\equ#1{Equation (\ref{equ:#1})}

\def\tab#1{Table \ref{tab:#1}}

\def\subsecs#1{\S\ref{subsec:#1}} 

\def\bestviewed{\textit{Best viewed in colour.}}

\newcommand*{\Scale}[2][4]{\scalebox{#1}{$#2$}}%
\newcommand*{\Resize}[2]{\resizebox{#1}{!}{$#2$}}%

\definecolor{Gray}{gray}{0.85}
\definecolor{LightCyan}{rgb}{0.88,1,1}
\usepackage{tcolorbox}

\newtcbox{\entoure}[1][orange]{on line,
arc=1pt,colback=#1!30!white,colframe=#1!100!black,
before upper={\rule[-0pt]{0pt}{5pt}},boxrule=.5pt,
boxsep=0pt,left=1pt,right=1pt,top=0pt,bottom=0pt}

\usepackage[pagebackref=true,breaklinks=true,letterpaper=true,colorlinks,bookmarks=false]{hyperref}

\iccvfinalcopy 

\def\iccvPaperID{2861} 

\ificcvfinal\fi
\begin{document}

\title{Spatio-temporal Prompting Network for Robust Video Feature Extraction}

\author{
Guanxiong Sun\textsuperscript{\rm 1, \rm 2}, Chi Wang \textsuperscript{\rm 1}, Zhaoyu Zhang \textsuperscript{\rm 1}, Jiankang Deng \textsuperscript{\rm 2, \rm 3}, Stefanos Zafeiriou
\textsuperscript{\rm 3}, Yang Hua\textsuperscript{\rm 1}
}
\affil{
    \textsuperscript{\rm 1}Queen's University Belfast \quad
    \textsuperscript{\rm 2}Huawei UKRD \quad
    \textsuperscript{\rm 3}Imperial College London \\
    \{gsun02, cwang38, zzhang55, y.hua\}@qub.ac.uk, 
    \{j.deng16, s.zafeiriou\}@imperial.ac.uk
}
\maketitle

\begin{abstract}
Frame quality deterioration is one of the main challenges in the field of video understanding. To compensate for the information loss caused by deteriorated frames, recent approaches exploit transformer-based integration modules to obtain spatio-temporal information.
However, these integration modules are heavy and complex.
Furthermore, each integration module is specifically tailored for its target task, making it difficult to generalise to multiple tasks. 
In this paper, we present a neat and unified framework, called Spatio-Temporal Prompting Network (STPN). It can efficiently extract robust and accurate video features by dynamically adjusting the input features in the backbone network. Specifically, STPN predicts several video prompts containing spatio-temporal information of neighbour frames.
Then, these video prompts are prepended to the patch embeddings of the current frame as the updated input for video feature extraction. 
Moreover, STPN is easy to generalise to various video tasks because it does not contain task-specific modules.
Without bells and whistles, STPN achieves state-of-the-art performance on three widely-used datasets for different video understanding tasks, i.e., ImageNetVID for video object detection, YouTubeVIS for video instance segmentation, and GOT-10k for visual object tracking. Code is available at \url{https://github.com/guanxiongsun/vfe.pytorch}
\end{abstract}

\section{Introduction}
\label{sec:intro}

\begin{figure}
\begin{center}
\includegraphics[width = \columnwidth]{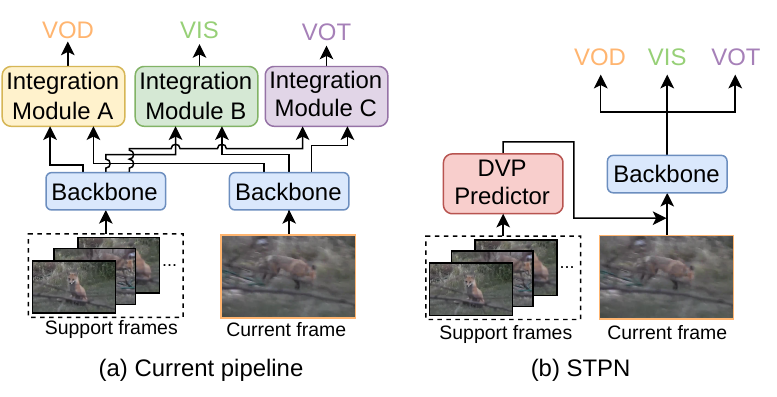}
\end{center}
\caption{
Comparisons between pipelines of (a) existing methods and (b) the proposed Spatio-temporal Prompting Network (STPN). Existing methods introduce complex and \textbf{task-specific} integration modules (\entoure[Goldenrod]{\ \ \ \  } \entoure[YellowGreen]{\ \ \ \   } \entoure[Orchid]{\ \ \  \  }) \textit{after} backbone networks (\entoure[RoyalBlue]{\ \ \ \   }). In contrast, STPN is a \textbf{unified} framework for multiple tasks. A lightweight dynamic video prompt (DVP) predictor (\entoure[Maroon]{\ \ \ \   }) generates a set of DVPs to adjust input \textit{before} backbone networks (\entoure[RoyalBlue]{\ \ \ \   }). \bestviewed
}
\label{fig:pipeline}
\end{figure}

Video understanding is a fundamental research direction in the field of computer vision. It plays an important role in many real-world applications, such as autonomous driving \cite{autonomous_driving1,autonomous_driving2}, video surveillance \cite{surveillance1,surveillance2}, and sports analysis \cite{sport,sport1m}.
However, a significant challenge in this field is the deterioration of video frames due to motion blur, occlusion, and deformation, which makes it difficult to extract relevant information.

To overcome this challenge, inspired by the great success of transformers in many computer vision tasks \cite{attention,nonlocal,vit,rn,detr}, researchers explore various transformer-based integration modules to alleviate the information loss on the deteriorated video frames.
For example, in video object detection, transformer-based feature aggregation methods \cite{rdn,mega,selsa,mamba} are investigated to enhance the feature of proposals \cite{fasterrcnn} in the detection head. In video instance segmentation, deteriorated frames usually cause wrong instance associations in a sequence, so 3D mask decoders \cite{vistr,ifc,mask2formervis} are introduced to learn associations of instance masks in an end-to-end manner. For visual object tracking, traditional correlation filters \cite{correlation,cfnet} meet their limitations in severely degraded frames. Therefore, transformer-based integration modules \cite{mixformer,stark,transt,treg} are proposed to better capture complicated correlations between the template region and the search region. To conclude, the summarised pipeline of recent methods is shown in \fig{pipeline} (a). A backbone network extracts spatially-only features from the current frame and the support frames in the same video. Then, different integration modules integrate spatial-only features on multiple frames to obtain spatio-temporal features. Finally, the spatio-temporal features are used for the targeted video understanding task.

While the current pipeline achieves good performance, there are still two major limitations of transformer-based integration modules.
Firstly, these integration modules are involved as an extra component after the backbone network, leading to increased complexity and additional computational costs. Secondly, each integration module is tailored specifically for the target task and thus cannot be generalised to multiple video tasks. 
Hence, we put forward the following question: can we remove the complex integration modules and directly obtain spatio-temporal information in backbone networks?

To answer this question, inspired by recent prompting techniques \cite{prompt_efficient_video_understanding,vpt,prompt}, in this paper, we present a neat and unified framework, called Spatio-Temporal Prompting Network (STPN). 
Instead of using complex integration modules, STPN simplifies the current pipeline for video understanding by introducing spatio-temporal information into the backbone network, as shown in \fig{pipeline} (b). 
Specifically, given a video frame, we propose a dynamic video prompt (DVP) predictor to generate several video prompts according to support frames. Then, the predicted DVPs are prepended to the patch embeddings of the current frame as the updated input. Finally, a vision transformer backbone network extracts video features using the updated input for future video understanding tasks. 
It is worth noting that the DVP predictor is a lightweight structure and introduces only a small number of extra parameters, e.g., 0.11M. Moreover, STPN can be easily adapted to different video understanding tasks, since it does not contain task-specific modules, and all modifications happen before the backbone network.

In summary, our key contributions are: (1) We present the Spatio-Temporal Prompting Network (STPN) that can extract robust video features on deteriorated video frames. STPN simplifies the current pipeline for video understanding and is easy to generalise to different video understanding tasks.
(2) To the best of our knowledge, we are the first to explore promoting techniques for robust video feature extraction on the task of video object detection (VOD), video instance segmentation (VIS), and visual object tracking (VOT). 
(3) Without bells and whistles, STPN achieves state-of-the-art performance on three widely-used video understanding benchmarks, i.e., ImageNet-VID \cite{imagenet} for VOD, YouTube-VIS \cite{youtubevis} for VIS, and GOT-10k \cite{got10k} for VOT.

\section{Related Work}
\label{sec:related}

In this section, we first review the state-of-the-art (SOTA) methods for related video understanding tasks. Then, we introduce the visual prompting techniques.

\noindent\textbf{Video Object Detection (VOD).} 
SOTA VOD methods use attention modules to aggregate features of support frames onto the current frame and thus enhance the feature quality of the current frame. For example, RDN \cite{rdn} and SELSA \cite{selsa} utilise relation modules to aggregate proposal features from local frames
and global frames, respectively. 
These methods use sample memory architectures to store features on support frames, e.g., a sliding window or a memory queue, which limit the ``diversity" of support features.
Therefore, MAMBA \cite{mamba} proposes a memory bank architecture that can partially update and sample support features and thus achieve a good speed-accuracy trade-off. More recent approaches TransVOD \cite{transvod} and TDViT \cite{tdvit} introduce transformers into the field and further improve the performance of VOD. 
\leavevmode\newline

\noindent\textbf{Video Instance Segmentation (VIS).} 
The goal of VIS is to simultaneously segment and track all object instances in the video sequence. There are two categories of approaches: Firstly, \textit{per-frame} approaches conduct instance segmentation on each frame independently and then associate instance masks by post-processing. MaskTrackRCNN \cite{youtubevis}
adds a tracking prediction head on MaskRCNN \cite{maskrcnn} for instance association.
More recently, some approaches \cite{idol, minvis, vita} use transformers \cite{attention,detr} to reason the relationships between instances.
Secondly, \textit{per-clip} approaches \cite{vistr,seqformer,mask2formervis,ifc} directly predict 3D masks for each instance in a video. VisTR \cite{vistr} proposes a transformer encoder-decoder structure to obtain spatio-temporal features, which are then used in an instance sequence matching module to get object tracks. IFC \cite{ifc} uses memory tokens inside of the transformer encoder module to transfer temporal information between input frames.
\leavevmode\newline

\noindent\textbf{Visual Object Tracking (VOT).}
Object tracking is one of the most fundamental research topics in computer vision and has a long history. 
Siamese-based methods \cite{siamese_fc, siamrpn,siamrpn++, siammask, distractor} consider tracking as a template-matching problem by extracting correlation feature maps of the target and the search region via two-head siamese networks. Recently,
transformers have been introduced to perform attention-based correspondence modelling and these methods \cite{transt,tmt,treg,stark} achieve SOTA performance.
Among these trackers, MixFormer \cite{mixformer} is a unified framework established solely on transformers for feature extraction, correspondence modelling, and result generation. 
\leavevmode\newline

\noindent\textbf{Prompting.} The concept of prompting is originally proposed in natural language processing (NLP). 
Prompting refers to designing instructions and prepending them to the input so that the pre-trained language models \cite{gpt,bert,gpt3} can ``understand" the task. 
Recently, prompting techniques have been explored in many computer vision tasks. A major trend \cite{xclip,stadapter,prompt_efficient_video_understanding} is to transfer pre-trained vision-language models \cite{vilbert, clip,zstig} to the task of video recognition, e.g., video classification and action recognition. 
Another popular research direction is parameter-efficient transfer learning via prompt tuning. For example, VPT \cite{vpt} and UPT \cite{upt} tune a small number of parameters in the input space of a model and can achieve good performance comparable to the full fine-tuning scheme on a variety of recognition tasks.


\begin{table}[]
\begin{center}
\resizebox{\columnwidth}{!}{%
\begin{tabular}{l|c}
\toprule
Meaning                           & Symbol  \\
\midrule
The $i$-th transformer layer                            & $\mathbf{L}_i$  \\
Collection of the output embeddings from $\mathbf{L}_i$ & $\mathbf{E}^i $ \\
Number of embeddings in $\mathbf{E}^i $                 & $n_i$           \\
Dimension of embeddings in $\mathbf{E}^i $              & $d_i$           \\
Collection of dynamic video prompts                            & $\mathbf{P}$    \\
Number of dynamic video prompts                         & $N_P$   \\
\bottomrule
\end{tabular}%
}%
\end{center}
\caption{List of notations.}
\label{tab:notations}
\end{table}

\section{Preliminaries}
\label{sec:preliminaries}
In this section, we define notations necessary for subsequent demonstrations.
$I_t$ denotes the current frame 
and $t$ is the current time step in the video. $I_{sup}$ denotes the support frames used to predict video prompts and aid in the feature extraction process. Specifically, $I_{sup} = \{I_\tau\}_{t-S(K/2)}^{t+S(K/2)}$, where $K$ denotes the number of support frames (assumed to be an even number) and $S$ denotes the frame interval in the time dimension between support frames. 

Transformers encoders \cite{vit,swin,cvt} are used to extract feature embeddings from video frames. A transformer encoder contains a patch embedding layer and $L$ transformer layers. The input frame is first divided into $n$ non-overlapping patches $\{\mathbf{x}_j \in \mathbb{R}^{3 \times h \times w}| j \in \mathbb{N}, 1 \leq j \leq n\}$, where $h$ and $w$ are the height and width of the patches. Each patch is then embedded into patch embeddings, denoted as $\mathbf{E} \in \mathbb{R}^{n_0\times d_0}$. Other notations are summarised in \tab{notations}.

\begin{figure*}
\begin{center}
  \includegraphics[width = \textwidth]{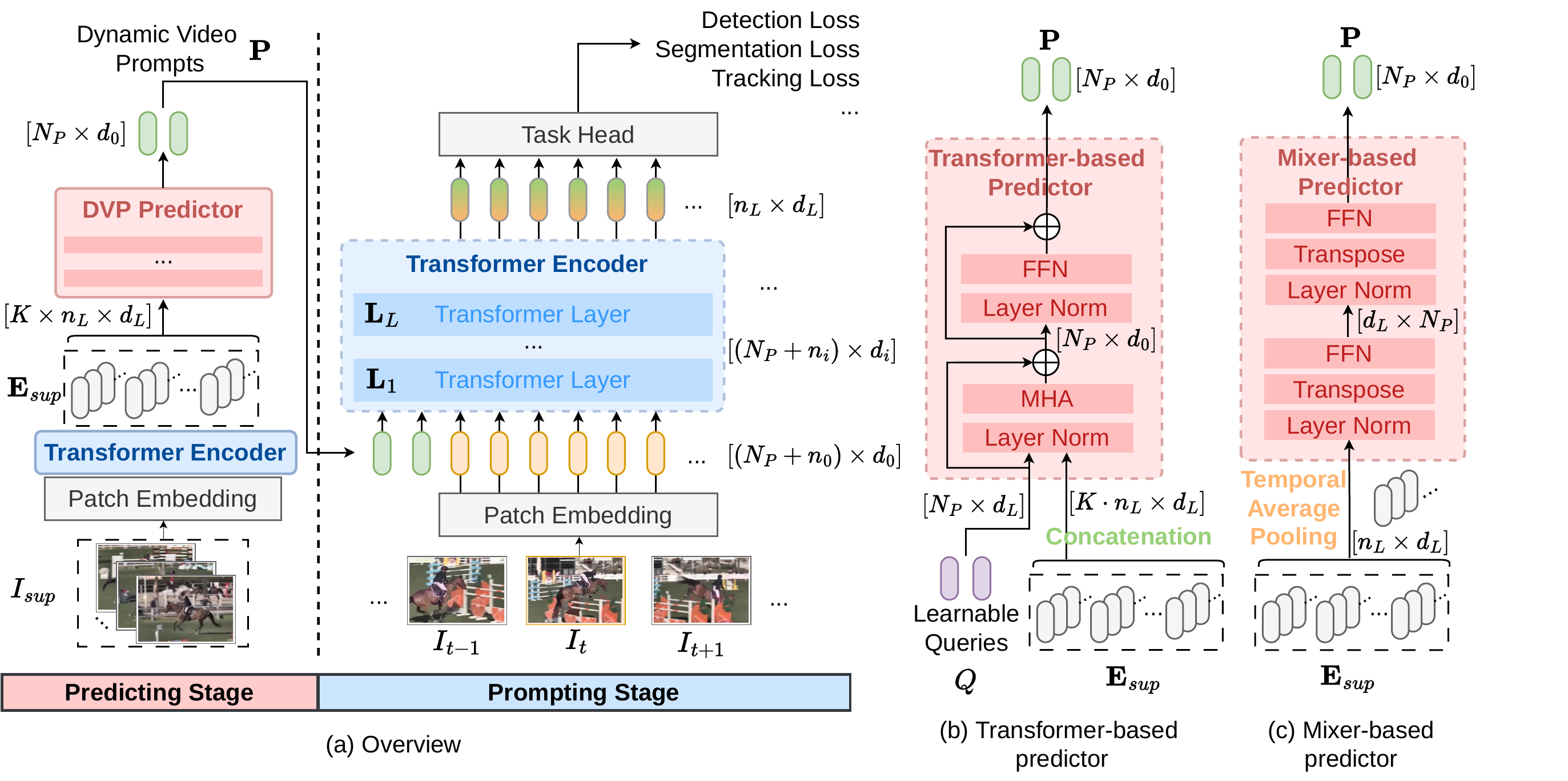}
\end{center}
\caption{
(a) An overview of our approach. In the predicting stage, a set of dynamic visual prompts (DVPs) $\mathbf{P}$ is generated by the DVP predictor that takes support embeddings $\mathbf{E}_{sup}$ on support frames $I_{sup}$ as the input. Then, in the prompting stage, predicted DVPs are prepended with the patch embeddings of the current frame to extract spatio-temporal embeddings via a transformer encoder which contains $L$ transformer layers. Finally, different task heads take the spatio-temporal embeddings and output final results for various general video tasks, e.g., video object detection, video instance segmentation, and visual object tracking. (b) Details of the transformer-based predictor. (c) Details of the Mixer-based predictor. 
}
\label{fig:overview}
\end{figure*}

\section{Methodology}
\label{sec:method}

\subsection{Overview}
\label{subsec:overview}

The overview of our framework is shown as in \fig{overview} (a). There are two stages in the framework: the predicting stage and the prompting stage. The goal of the predicting stage is to generate a set of prompts that vary according to spatio-temporal information. 
We first select $K$ support frames around $I_t$ and then pass the support frames $I_{sup}$ to the patch embedding layer and the transformer encoder to extract image embeddings, called support embeddings. The support embeddings are then passed to the dynamic video prompt (DVP) predictor network as the input. The DVP predictor outputs $N_P$ dynamic video prompts, denoted as $\mathbf{P}$, which contain the spatio-temporal information of the current video. Details of the DVP predictor network are illustrated in \subsecs{predicting}.

The second stage is the prompting stage whose goal is to extract spatio-temporal features of the current frame $I_t$ using $\mathbf{P}$. Specifically, $\mathbf{P}$ is prepended to the patch embeddings of the current frame $I_t$. Then, the mixed embeddings are passed into the transformer encoder to produce the spatio-temporal embeddings. Finally, the spatio-temporal embeddings are used for different video understanding tasks, e.g., video object detection, video instance segmentation and visual object tracking. More details of the prompting stage are presented in \subsecs{prompting}.

\subsection{Predicting Stage}
\label{subsec:predicting}
Given the current frame $I_t$, we first sample $K$ support frames around $I_t$ as support frames. The support frames are then passed into the patch embedding layer and the transformer encoder:
\begin{align}
{\mathbf{E}^i_\tau}  =\mathbf{L}^i(\mathbf{E}^{i-1}_{\tau}) \quad {\mathbf{E}^i_\tau} \in \mathbb{R}^{n_i\times d_i}, i=1,2, \ldots, L.
\end{align}
Therefore, the collection of support embeddings is denoted as:
$
\mathbf{E}_{sup} = \{ \mathbf{E^{L}_\tau} \in \mathbb{R}^{n_L \times d_L} | \tau \in \mathcal{T}\}, 
$
where $\mathcal{T}$ is the collection of sampled time steps around the current frame, $\mathcal{T} = \left[ t-S(K/2), ..., t-S, t+S, ..., t+S(K/2)\right]$. For example, if $K=$ 6 and $S=$ 2, three support frames are sampled from the past of the current time step $t$, and three are sampled from the future, with a time interval of 2 between each sampled frame in both directions.
Given $\mathbf{E_{sup}}$, we design two predicting networks, i.e., transformer-based predictor and mixer-based predictor, to obtain dynamic video prompts $P \in \mathbb{R}^{N_P \times d}$. 
\leavevmode\newline

\noindent\textbf{Transformer-based Predictor.} This design is inspired by the object decoder in DETR \cite{detr} where a set of randomly initialised object queries are used to retrieve object features on the image embeddings. As depicted in \fig{overview} (b), we introduce a set of learnable prompt queries $Q \in \mathbb{R}^{N_P \times d_L}$ to gather spatio-temporal information from the support embeddings. 
Specifically, a  multi-head attention (MHA) \cite{attention} module takes the support embeddings and the prompt queries as input, and produces the aggregated embeddings. The process is formulated as follows:
\begin{align}
K &= \text{Concat}(\mathbf{E_{sup}}) \quad K \in \mathbb{R}^{K \cdot n_L \times d_L}, \\
\hat{Q} &= \text{MHA}(\text{LN}(Q), \text{LN}(K)) + Q \quad \hat{Q} \in \mathbb{R}^{N_P \times d_L},
\end{align}
where $\text{Concat}(\cdot)$ indicates the concatenation operation and LN denotes the LayerNorm layer. 
The output embeddings are then added back to the prompt queries and passed to the LayerNorm layer. The normalised output embeddings are further transformed by a simple feed-forward network (FFN) with a residual connection that makes the dynamic video prompts:
\begin{align}
    \mathbf{P} = \text{FFN}(\text{LN}(\hat{Q})) + \text{LN}(\hat{Q}) \quad \mathbf{P} \in \mathbb{R}^{N_P \times d}.
\end{align}

\noindent\textbf{Mixer-based Predictor.} 
Inspired by the MLP-Mixer \cite{mlpmixer} (shortened as Mixer), we design a Mixer-based predictor to transform support embedding into dynamic video prompts. The details of a Mixer-based predictor are shown in \fig{overview} (c). Unlike the transformer-based predictor whose first step is to concatenate all support embeddings, we conduct average pooling in the temporal dimension to obtain the average support embeddings $\overline{\mathbf{E}}_{sup}$ as follows:
\begin{align}
\overline{\mathbf{E}}_{sup} = \text{AvgPool}(\mathbf{E}_{sup}) \quad \overline{\mathbf{E}}_{sup} \in \mathbb{R}^{n_L \times d_L},
\end{align}
where AvgPool denotes the average pooling across the time dimension. Then, the average support embedding is normalised by a LayerNorm layer and further transposed and passed into a feed-forward network (FFN) to obtain the hidden feature $h$ with $N_P$ channels. Finally, the hidden feature $h$ is used to generate $N_P$ dynamic video prompts with $d$ channels. These two steps can be formulated as follows:
\begin{align}
h &= \text{FFN}((\text{LN}(\overline{\mathbf{E}}_{sup}))^{T}) \quad h \in \mathbb{R}^{d_L \times N_P}, \\
\mathbf{P} &= \text{FFN}((\text{LN}(h)^{T}) \quad \mathbf{P} \in \mathbb{R}^{N_P \times d},
\end{align}
where LN and $(\cdot)^T$ denote the LayerNorm layer \cite{layernorm} and the transpose layer, respectively.


\subsection{Prompting Stage}
\label{subsec:prompting}

Given the predicted dynamic video prompts $\mathbf{P}$, we concatenate $\mathbf{P}$ 
together with the patch embeddings of the current frame $\mathbf{E}_t$ as the input of the transformer encoder. The details of this process are shown in the prompting stage (right part) of \fig{overview} (a). More concretely, the process of applying $\mathbf{P}$ and extracting the spatio-temporal feature of the current frame $I_t$ can be formulated as:
\begin{align}
    \left[\mathbf{Z}^1_t, \mathbf{E}^1_t\right] &= \mathbf{L}^1(\text{Concat}(\mathbf{P}, \mathbf{E}_t)) \\
    \left[\mathbf{Z}^i_t, \mathbf{E}^i_t\right] &= \mathbf{L}^i(\text{Concat}(\mathbf{Z}^{i-1}, \mathbf{E}^{i-1}_t)) \quad i=2,3,...,L \label{equ:extract} \\
    y_t  &= \text{Head}(\mathbf{E}^L_t),
\end{align}
where $\text{Concat}(\cdot)$ indicates the concatenation operation and $\mathbf{Z}^i_t \in \mathbb{R}^{N_P \times d_i}$ represents the extra embeddings produced by adding $\mathbf{P}$ in the input embeddings. The weights of the transformer encoder are shared in the predicting and prompting stages.
Finally, the output embeddings $\mathbf{E}^L_t$ are passed into the head network for different video tasks.
\leavevmode\newline

\begin{figure}
\begin{center}
\includegraphics[width = 0.9\columnwidth]{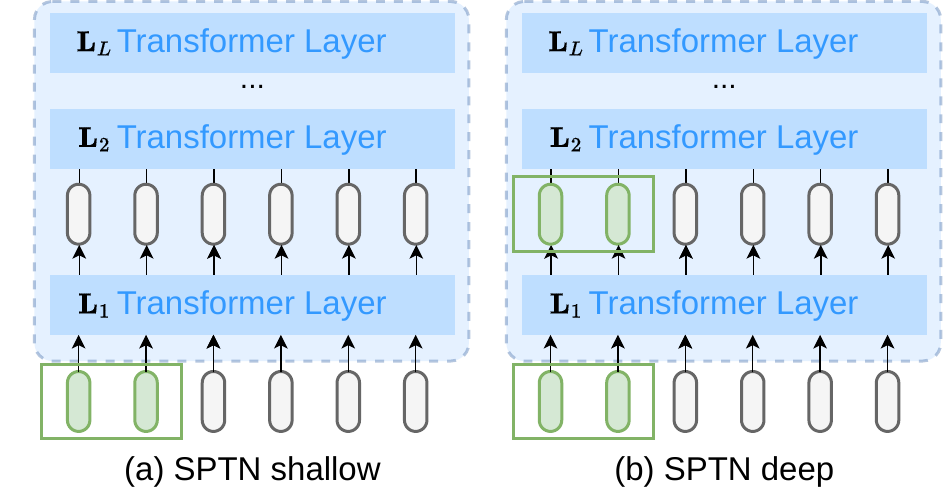}
\end{center}
\caption{
A comparison between the (a) STPN shallow and the (b) STPN deep. The green rounded rectangles denote predicted DVPs. In (a), one set of DVPs is prepended to the patch embeddings before the \textbf{first} layer of the transformer encoder. In (b), $L$ sets of DVPs are predicted and then prepended to the input embeddings of \textbf{all} $L$ transformer layers in the transformer encoder. 
}
\label{fig:deep_shallow}
\end{figure}

\noindent\textbf{STPN Shallow $v.s.$ STPN Deep.}
Following the protocols in VPT \cite{vpt}, we design two STPN variants, i.e., STPN shallow $v.s.$ STPN deep, according to how many sets of dynamic video prompts (DVPs) are generated to adjust the transformer encoder. For example, in the shallow version, one set of DVPs is predicted and prepended only before the first transformer layer. In the deep version, $L$ sets of DVPs are predicted via L separate FC layers, and each set of DVPs is prepended to the corresponding transformer layer. Formally, the DVPs of the $i$-th layer are denoted as $\mathbf{P}^i \in \mathbb{R}^{N_P \times d_i}$ and the feature extraction process \equ{extract} is changed to:
\begin{align}
\Scale[0.95]{
    \left[ \ldots , \mathbf{E}^i_t \right] = \mathbf{L}^i(\text{Concat}(\mathbf{P}^{i-1}, \mathbf{E}^{i-1}_t)) \quad i=2,3,...,L. }
\end{align}
The difference between the shallow and deep variants is shown in \fig{deep_shallow}.

\section{Experiments}
\label{sec:experiment}

We first conduct experiments and compare SPTN with other state-of-the-art methods on three general video understanding tasks, i.e., video object detection (VOD), video instance segmentation (VIS) and visual object tracking (VOT). Then, in \subsecs{ablation}, we conduct ablation studies on how different design choices affect the overall performance, which helps to gain a deeper understanding of SPTN.

\subsection{Settings}

\noindent\textbf{Datasets.} For VOD, the ImageNetVID dataset \cite{imagenet} is used. It contains 3,862 training and 555 validation videos. Following previous approaches  \cite{mamba,selsa,fgfa}, we add overlapped 30 classes of the ImageNet DET dataset into the $train$ set. Specifically, we sample 15 frames from each video in the VID $train$ and at most 2,000 images per class from the DET $train$. For VIS, we train our models and report results on the YouTube VIS 2019 dataset \cite{youtubevis}. The dataset contains 2,238 training, 302 validation, and 343 test videos, which covers 40 object categories. For VOT, we choose the GOT10k \cite{got10k} dataset, because it is a large-scale dataset (10,000 training videos and 180 test videos) with more than 1.5 million manually labelled bounding boxes, enabling stable training and evaluation of trackers.
\leavevmode\newline

\noindent\textbf{Evaluation Metrics.}
For video object detection and video instance segmentation, we report detailed results in the COCO \cite{coco} evaluation format. Specifically, we report the average precision (AP) metric which computes the average precision over ten IoU thresholds $\left[0.5: 0.05: 0.95\right]$ for all categories. Meanwhile, the COCO evaluation contains other important metrics, e.g., AP$_{50}$ and AP$_{75}$ that are calculated at IoU thresholds of 0.50 and 0.75, and AP$_S$, AP$_{M}$,  AP$_{L}$ that are calculated on different object scales, i.e., small, medium and large. The metrics for detection and segmentation are noted with ${box}$ and ${mask}$, respectively. For visual object tracking, we report the widely used average overlap (AO) and the success rate (SR). AO denotes the average overlaps between all ground truth and estimated bounding boxes, while SR measures the percentage of successfully tracked frames where the overlaps exceed a threshold, e.g., SR$_{0.5}$ and SR$_{0.75}$.\leavevmode\newline

\noindent\textbf{Transformer Encoders.} Our method is compatible with different transformer backbones. For demonstration, we conduct experiments on two well-known transformer backbones, i.e., Swin \cite{swin} and CvT \cite{cvt}.
Swin is used for video object detection and video instance segmentation. CvT is used for visual object tracking. The details of the backbones are illustrated in the supplementary material.\leavevmode\newline

\noindent\textbf{Training and Inference.}
All reported results are obtained with Python 3.7 and PyTorch 1.8.1 on Tesla V100 GPUs. We follow the protocols in TransVOD \cite{transvod}, Mask2Former \cite{mask2former}, and MixFormer \cite{mixformer} to set the hyper-parameters, e.g., learning rate, data augmentations, optimiser, etc., for three tasks, respectively. More details about the hyper-parameter settings are described in the supplementary material.

\begin{table*}[t]
\begin{center}
\begin{tabular}{lllllllll}
\toprule
Model                 & Base Detector   & Backbone   & AP   & AP50 & AP75          & AP$_S$  & AP$_M$  & AP$_L$  \\ \midrule
CHP \cite{chp}        & CenterNet       & R101       & -    & 76.7 & -             & -    & -    & -    \\
EOVOD    \cite{eovod} & FCOS            & R101       & 54.1 & 79.8 & 59.5          & 10.5 & 28.3 & 60.1 \\
RDN   \cite{rdn}      & FasterRCNN      & R101       & 53.4 & 81.8 & 60.1          & 8.5  & 27.4 & 59.6 \\
TransVOD \cite{transvod}            & Deformable DETR & R101       & -    & 82.0 & -             & -    & -    & -    \\
MEGA  \cite{mega}     & FasterRCNN      & R101       & 53.2 & 82.9 & 59.2          & 9.1  & 29.4 & 59.1 \\
TDViT    \cite{tdvit} & FasterRCNN & SwinT & 50.9          & 79.9          & 55.7 & 9.1           & 26.9          & 57.2          \\
SELSA*  \cite{selsa}              & FasterRCNN      & SwinT      & 52.3 & 82.3 & 58.2          & 11.4 & 28.8 & 57.9 \\
TransVOD  \cite{transvod}            & Deformable DETR & SwinT      & -    & 83.7 & -             & -    & -    & -    \\
\midrule
STPN                   & FasterRCNN      & SwinT      & 60.6 & 85.0 & \textbf{68.8} & 12.1 & 33.7 & 66.8 \\
STPN         & SELSA      & SwinT & \textbf{60.9} & \textbf{86.5} & 68.6 & \textbf{13.3} & \textbf{34.8} & \textbf{67.1} \\
\midrule
MEGA   \cite{mega}               & FasterRCNN      & ResNeXt101 & -    & 84.1 & -             & -    & -    & -    \\
MAMBA    \cite{mamba} & FasterRCNN      & ResNeXt101 & -    & 86.7 & -             & -    & -    & -    \\
TransVOD   \cite{transvod}           & Deformable DETR & SwinB      & -    & 90.0 & -             & -    & -    & -    \\
\midrule
STPN                   & FasterRCNN      & SwinB      & 63.6   & 86.1   & 71.6             &  12.1    &  33.8    &  70.3    \\
STPN         & SELSA           & SwinB      &   \textbf{65.2 }     &   \textbf{90.6 }   &  \textbf{73.5}    & \textbf{14.1}     &  \textbf{37.7}  & \textbf{71.8 }   \\
\bottomrule
\end{tabular}
\end{center}
\caption{Comparison with state-of-the-art methods on ImageNet VID for video object detection. $\ast$ represents our re-implementation.}
\label{tab:sota_vod}
\end{table*}

\begin{table}[]
\begin{center}
\resizebox{\columnwidth}{!}{%
\begin{tabular}{llllll}
\toprule
Model                             & Backbone & AP   & AP50 & AP75 & FPS  \\
\midrule
VisTR  \cite{vistr}               & R101     & 40.1 & 64.0 & 45.0 & 57.7 \\
CrossVIS                          & R101     & 36.6 & 57.3 & 39.7 & 35.6 \\
MaskProp                          & R101     & 42.5 & -    & 45.6 & -    \\
SeqFormer                         & R101     & 49.0 & 71.1 & 55.7 & 64.6 \\
IFC   \cite{ifc}                  & R101     & 44.6 & 69.2 & 49.5 & \textbf{89.4} \\
IDOL \cite{idol} &R101 &50.1 &73.1 &56.1 & 26.0\\
VITA \cite{vita} &R101 &51.9 &75.4 &57.0 & - \\
M2F-VIS \cite{mask2formervis} & SwinT    & 51.5 & 75.0 & 56.5 & -    \\
MinVIS   \cite{minvis}        & SwinT    & 51.9 &  75.1   &  58.2    &  27.1    \\
\midrule
STPN$_{MinVIS}$       & SwinT    & \textbf{54.8} &  \textbf{78.7}    &  \textbf{61.0}    & 26.5     \\
\bottomrule
\end{tabular}
}
\end{center}
\caption{Comparison with state-of-the-art methods for video instance segmentation on the YouTube VIS 2019 dataset.}
\label{tab:sota_vis}
\end{table}

\subsection{Comparisons with SOTA Methods}
\label{subsec:sota}
\noindent\textbf{Video Object Detection.} \tab{sota_vod} shows the results of STPN and state-of-the-art methods. Firstly, in the AP50 metric, using a simple FasterRCNN \cite{fasterrcnn} base detector with the Swin-T \cite{swin} backbone, STPN achieves 85.0\% of AP50, which makes a 1.3\% improvement over the best competitor TransVOD \cite{transvod}. Furthermore, when changing the FasterRCNN base detector to a multi-frame aggregation method SELSA \cite{selsa}, STPN increases further to 86.5\% of AP50. When using a more strict AP metric, STPN with FasterRCNN and SELSA achieve remarkable improvements of +6.5/6.9\% over the best competitor EOVOD \cite{eovod}, respectively. Finally, when changing to stronger backbones, STPN with the Swin-B backbone achieves further improved performance of 65.2\% AP. Compared to TransVOD, STPN with SELSA makes a 0.6\% improvement in AP50.\leavevmode\newline

\begin{table}[]
\begin{center}
\Resize{\columnwidth}{
\begin{tabular}{lllll}
\toprule
Model         & Backbone & AO   & SR0.5 & SR0.75 \\
\midrule
SiamRPN++ \cite{siamrpn++}  &R101   &51.7 &61.6 &32.5\\
SiamR-CNN \cite{siamrcnn}    & R101     & 64.9 & 72.8  & 59.7   \\
PrDiMP \cite{prdimp}    &R101   &63.4 &73.8 &54.3\\
TrDiMP \cite{trdimp}    &R101   &67.1 &77.7 &58.3 \\
TREG \cite{treg}          & R101     & 66.8 & 77.8  & 57.2   \\
TransT  \cite{transt}      & R101     & 67.1 & 76.8  & 60.9   \\
STARK \cite{stark}        & R101     & 68.8 & 78.1  & 64.1   \\
MixFormer\cite{mixformer} & CVT-22K      & 70.7 & 80.0  & 67.8   \\
MixFormer \cite{mixformer}  & CVT-1K      & 71.2 & 79.9  & 65.8   \\
\midrule
STPN$_{MixFormer}$           & CVT-22K      & 71.8 & 81.6  & \textbf{69.0}   \\
STPN$_{MixFormer}$           & CVT-1K      & \textbf{72.7} & \textbf{81.9}  & 67.7  \\
\bottomrule
\end{tabular}%
}%
\end{center}
\caption{Comparison with state-of-the-art tracking methods on the GOT-10k dataset.}
\label{tab:sota_got}
\end{table}

\noindent\textbf{Video Instance Segmentation.}
We compare STPN with state-of-the-art methods in \tab{sota_vis}. Using the M2F-VIS \cite{mask2formervis} as the baseline segmentation method, STPN improves performance by large margins with 2.9/3.6\% of AP/AP50, respectively. Compared with SOTA methods with similar backbone complexities, for example, Swin-T or ResNet-101, STPN achieves the best accuracy. Specifically, STPN improves the performance of the best competitor VITA \cite{vita} by 2.9/3.3\% of AP/AP50, respectively. \leavevmode\newline

\begin{figure*}
    \begin{center}
    \includegraphics[width=\textwidth]{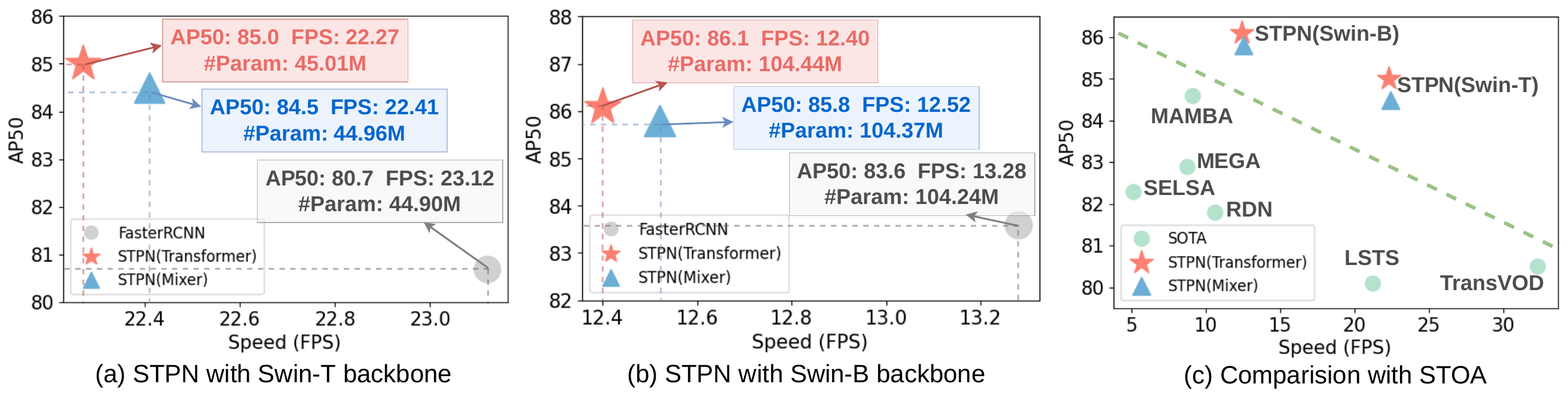}
    \end{center}
    \caption{Comparisons between the transformer-based and the Mixer-based DVP predictors.}
    \label{fig:speed_acc}
\end{figure*}

\begin{figure}
\begin{center}
\includegraphics[width = \columnwidth]{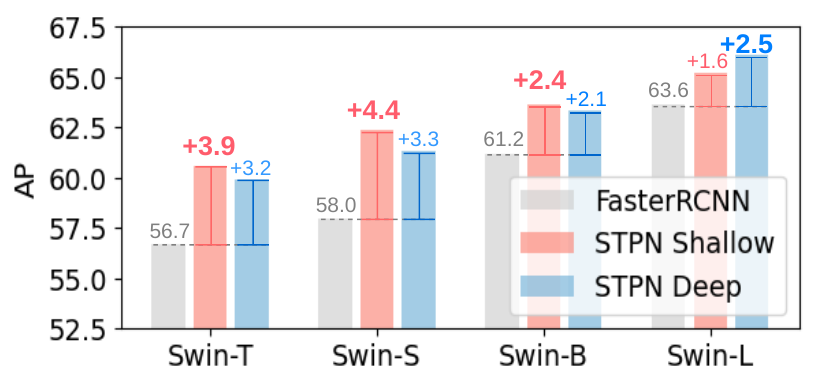}   
\end{center}
\caption{Comparisons of STPN deep and STPN shallow on transformer encoders with different scales.}
\label{fig:deep_shallow_exp}
\end{figure}

\noindent\textbf{Visual Object Tracking.} \tab{sota_got} compares STPN with state-of-the-art methods. We use MixFormer \cite{mixformer} as our baseline. Specifically, when CVT-22k and CVT-1k are used as the backbone, STPN improves the AO of MixFormer by +1.1\% and +1.5\%, respectively. STPN achieves the best performance compared to other SOTA methods on the GOT-10K dataset.

These experimental results demonstrate that STPN is easy to generalise to various video understanding tasks and achieves remarkable performance across all three different video tasks.

\subsection{Ablation Study}
\label{subsec:ablation}
We investigate the effect of different designs of our approach on the overall performance. All experiments in \subsecs{ablation} are conducted on the ImageNet VID dataset for video object detection. \leavevmode\newline

\noindent\textbf{DVP Predictor Designs.}
We compare the effects of DVP predictor designs on different baselines. \fig{speed_acc} (a) shows the results on the Swin-T backbone. The FasterRCNN baseline achieves 80.7\% AP50, runs at 23.12 FPS, and has 44.90M parameters. Using the transformer-based predictor and the Mixer-based predictor, STPN achieves 85.0/84.5\% AP50 and runs at 22.27/22.41 FPS and has 45.01/44.96M parameters, respectively. These results show that the transformer-based predictor can achieve better accuracy, while the Mixer-based predictor has less complexity and faster speed. Experiments on the Swin-B backbone are shown in \fig{speed_acc} (b) and the experimental results indicate the same conclusion. \leavevmode\newline

\noindent\textbf{Efficiency of STPN.} Compared with other state-of-the-art methods, STPN with both predictor designs actually achieves a significantly improved speed-accuracy trade-off, as shown in \fig{speed_acc} (c). These results demonstrate that the pipeline of STPN is very efficient and is not sensitive to different predictor designs. We choose the transformer-based predictor as our default setting since it achieves a slightly higher accuracy with a comparable fast speed. \leavevmode\newline

\noindent\textbf{STPN Shallow} \textit{v.s.} \textbf{STPN Deep.} 
As shown in the \fig{deep_shallow_exp}, we conduct experiments on Swin-T/S/B/L variants. The first finding is that both STPN shallow and STPN deep can continuously improve performance on all Swin variants, indicating that STPN has good compatibility with backbone scales.  Another finding which contradicts our intuition is that surprisingly STPN shallow achieves better performance than STPN deep on three Swin variants. We believe the reason is that STPN shallow has fewer parameters and thus is easier to optimise in the training process. In contrast, the Swin-L variant is much more complex and thus needs more adjustments in the intermediate transformer layers. By default, we use STPN shallow. \leavevmode\newline

\begin{table}[t]
\begin{center}
\resizebox{\columnwidth}{!}{%
\begin{tabular}{cc|cccccc}
\toprule
\multirow{2}{*}{(a)} & $S$   & 1    & 2    & 4             & 8             & 16   & 32   \\
                     & AP                   & 58.7 & 59.6 & 60.1          & \textbf{60.6} & 60.5 & 60.5  \\
\midrule
\multirow{3}{*}{(b)} & $K$   & 1    & 3    & 5             & 7             & 9    & 11   \\
                     & AP                   & 52.6 & 53.5 & 58.9          & \textbf{60.6} & 60.6 & 60.7 \\
                     & FPS                  & 22.4 & 22.4 & 22.3          & \textbf{22.3} & 21.7 & 20.5  \\
\midrule
\multirow{3}{*}{(c)} & $N_P$ & 3    & 5    & 7             & 9             & 11   & 13   \\
                     & AP                   & 60.2 & 60.3 & \textbf{60.6} & 60.6          & 60.5 & 60.6 \\
                     & FPS                  & 22.9 & 22.7 & \textbf{22.3} & 21.6          & 21.0 & 19.9 \\
\bottomrule
\end{tabular}%
}%
\end{center}
\caption{Effect of hyper-parameters in STPN: the temporal stride $S$, the number of support frames $K$, and the number of dynamic video prompts $N_P$. Different choices of hyper-parameters are listed in rows highlighted in grey.}
\label{tab:hyper}
\end{table}

\noindent\textbf{Support Frame Sampling.}
We use two hyper-parameters $S$ and $K$ to control the sampling of support frames. \tab{hyper} (a) and \tab{hyper} (b) show the effects of choosing different values of $S$ and $K$, respectively. Firstly, we fix the value of $K$ as 7, so 15 surrounding support frames are used in total. Then, we vary the temporal stride $S$ from 1 to 32. As shown in \tab{hyper} (a), S=8 achieves the best performance. Afterwards, with $S$ fixed as 8, we verify the effect of different $K$ with respect to both performance (AP) and speed (FPS). Specifically, when $K$=7, STPN achieves a good speed-accuracy trade-off with 60.6\% AP and 22.3 FPS. Therefore, we choose $S$ = 8 and $K$ = 7 as default. \leavevmode\newline

\noindent\textbf{Number of dynamic video prompts.} To explore the effect of the number of dynamic video prompts $N_P$, we show the performance and speed results by varying this number from 3 to 13 in \tab{hyper} (c). The best speed-accuracy trade-off is obtained when $N_P$ is $7$. In particular, once $N_P$ is greater than 7, AP is less affected by the change of $N_P$. Thus, we use 7 as the default value of $N_P$.

\begin{figure}[]
\begin{center}
\includegraphics[width = 0.9\columnwidth]{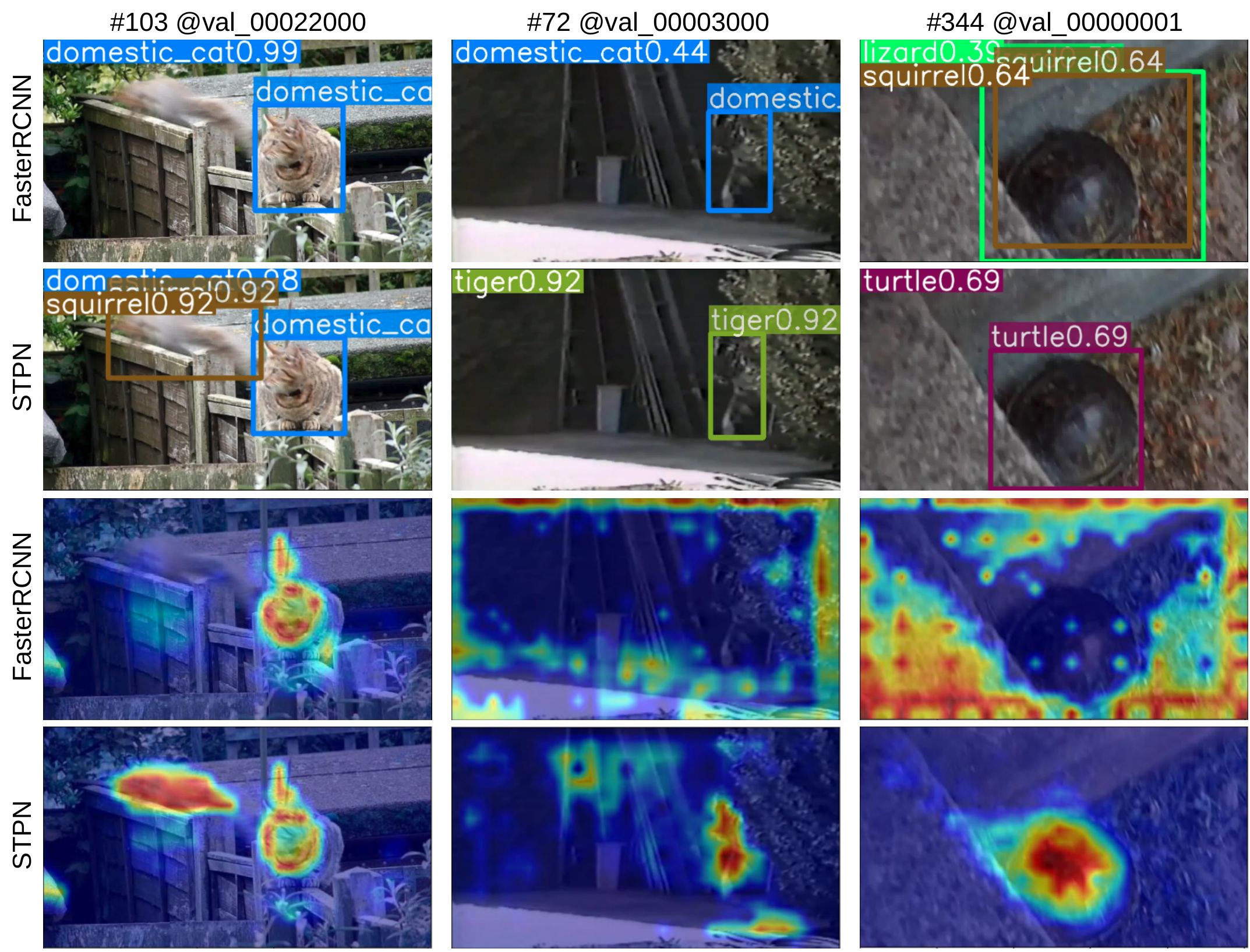}
\end{center}
\caption{Bounding boxes and Grad-CAM visualisations of FasterRCNN and STPN. Each column shows a video frame from the ImageNet VID validation set.}
\label{fig:gradcam}
\end{figure}

\begin{figure}[]
\begin{center}
\includegraphics[width = 0.9\columnwidth]{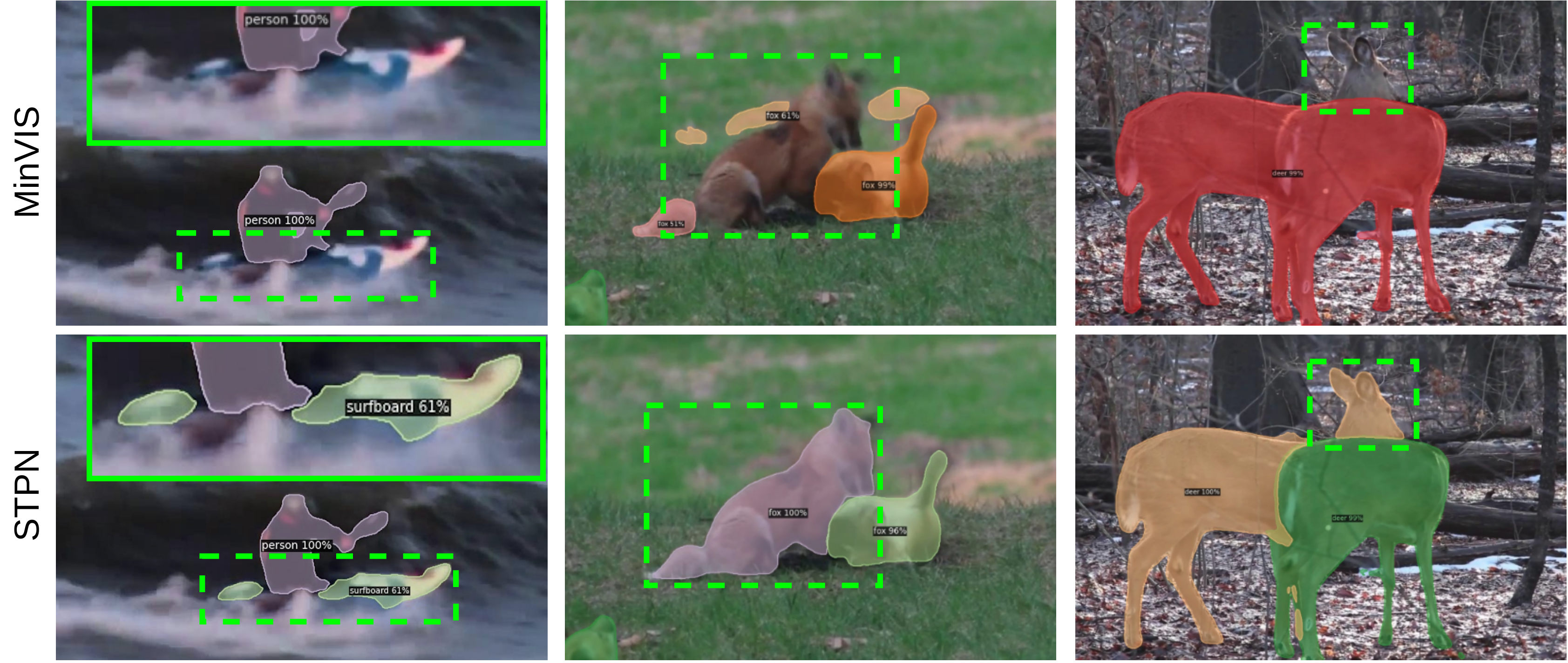}
\end{center}
\caption{Instance mask visualisations of MinVIS \cite{minvis} and STPN. Each column shows a video frame from the YouTube VIS dataset. MinVIS fails to generate accurate masks in low-quality frames.}
\label{fig:vis}
\end{figure}

\subsection{Qualitative Results}

\noindent\textbf{Grad-CAM} \cite{gradcam} is used to visualise the ``concentration" areas of the FasterRCNN baseline and STPN in some challenging scenarios, for example, motion blur, occlusion, and both. As shown in \fig{gradcam}, FasterRCNN either fails to detect the blurry object or generates false positive detections, while STPN can produce accurate detections. In the first column, the squirrel is very blurry because of its fast movement. As a result, FasterRCNN cannot detect the squirrel, and thus the Grad-CAM does not show much attention to the region of the squirrel. On the contrary, STPN successfully detects the squirrel, and Grad-CAM shows a good concentration map in the squirrel region. \leavevmode\newline

\noindent\textbf{Instance Mask Visualisation.} \fig{vis} shows the mask segmentation results of the state-of-the-art method MinVIS \cite{minvis} and the STPN in deteriorated frames. For example, in the first and the second columns, MinVIS fails to generate instance masks for the surfboard and the fox because of motion blur but our STPN can generate accurate instance masks in these frames. The third column shows a case of occlusion. The left deer is occluded by the right one, so the MinVIS fails to generate a mask in the head area of the left dear. Besides, it cannot distinguish the two deer and generate one instance mask. In contrast, STPN works well for generating the mask of the occluded deer and distinguishing the two instances of deer.\leavevmode\newline

\begin{figure}
\begin{center}
\includegraphics[width = 0.9\columnwidth]{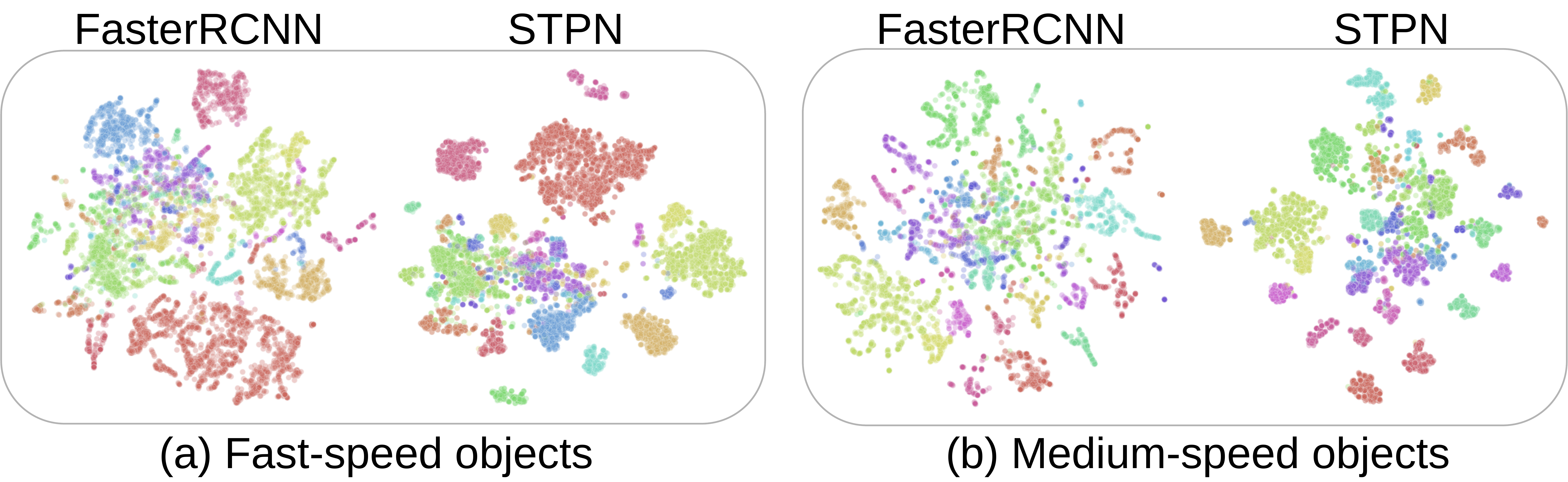}     
\end{center}
\caption{t-SNE visualisation of the object features produced by FasterRCNN and STPN on (a) fast speed and (b) medium speed objects in the ImageNet VID dataset.}
\label{fig:tsne}
\end{figure}

\noindent\textbf{t-SNE} \cite{tsne} is a statistical method for visualising high-dimensional data in 2D maps and is usually used to demonstrate the discriminative abilities of different models in classification tasks. We visualise the object features extracted by FasterRCNN and STPN using t-SNE. \fig{tsne} (a) and (b) show the results on the objects with the fast and medium moving speeds, respectively. We can see that STPN has better clustering results than the FasterRCNN baseline, demonstrating that STPN can improve the discriminative ability of features.
Details of how to define different moving speeds and how to generate object features are illustrated in the supplementary material.



\section{Conclusion}
To the best of our knowledge, we are the first to explore promoting techniques for robust video feature extraction. We present a neat and unified framework, named Sptio-Temporal Prompting Network (STPN). STPN simplifies the current pipeline for video understanding tasks. Moreover, STPN is easy to generalise to various video tasks. We conduct extensive experiments and report STPN's superior performance on three widely-used video benchmarks: ImageNet VID for video object detection, YouTube for video instance segmentation, and GOT-10K for visual object tracking. 
We hope our work can inform future research on robust video understanding.

\appendix
\section{Appendix}

We introduce detailed implementations of STPN.
We use Python 3.7 and PyTorch 1.8.1 \cite{pytorch}, and conduct experiments on NVIDIA Tesla V100-32GB GPUs.

\subsection{STPN on CVT}
\label{sec:cvt}

CVT \cite{cvt} is the default transformer encoder used in MixFormer \cite{mixformer} for visual object tracking. The architecture of CVT is slightly different from other transformer encoders \cite{swin,vit} mainly because CVT uses convolutional layers to generate down-scaled feature maps whereas other encoders are  down-scaled by reshaping and concatenating. As a result, the predicted DVP by STPN should also be compatible with the convolution-based downscale layers in CVT.

The convolution-based downscale layer in CVT is implemented by Conv2d layers with kernel size 3$\times$3 and stride 2. The predicted dynamic video prompts (DVPs) consist of $N_P$ embeddings. To enable Conv2d on DVPs, we increase $N_P$ from the default value of 5 to 9, so that we can reshape the DVPs to a 3$\times$3 feature map. The reshaped DVPs are then passed into the convolution-based downscale layer to generate down-scaled embeddings. Additionally, we add zero padding of size 2 on the reshaped DVPs before passing it to the Conv2d layer to ensure the output DVPs have the same size as the input. Therefore, the input size and the output size of DVPs are the same which is 3x3. Finally, the output DVPs are reshaped back to 9 embeddings and then prepended with down-scaled feature maps for the following transformer layer.


\subsection{Training and Inference}

Details of hyper-parameters we used for video object detection (VOD), video instance segmentation (VIS), and visual object tracking (VOT) are listed in the \tab{hyper}. For VOD and VIS, following the training protocols in FasterRCNN \cite{fasterrcnn} and MinVIS \cite{minvis}, the whole model is trained end-to-end in a single stage. In contrast, following the training protocol in MixFormer \cite{mixformer}, the training process is divided into two stages. The parameters in the score prediction head \cite{mixformer} are trained in the second stage. All other parameters including the DVP predictor and CVT backbone are trained in the first stage.

\subsection{Grad-CAM Details}

We use the EigenCAM \cite{eigencam} for visualising the class activation maps (CAM) for STPN on the task of video object detection. 
During the visualisation progress, two extra modifications are needed compared with the normal grad-cam visualisation for the task of image classification. 
Firstly, we need to formulate a customised ``reshape" transformation that integrates the stored activations in the FasterRCNN \cite{fasterrcnn} output features (from the feature pyramid network (FPN) \cite{fpn}). Specifically, we re-scale all feature levels of FPN to the same scale, the scale of 64$\times$ in our implementation. Secondly, we need to construct a ``target" function that generates CAMs optimised for specific bounding boxes, such as their score or their intersection over union with the original bounding boxes. More implementation details can be found in the  GitHub page \footnote{\href{https://github.com/jacobgil/pytorch-grad-cam/blob/master/tutorials/Class\%20Activation\%20Maps\%20for\%20Object\%20Detection\%20With\%20Faster\%20RCNN.ipynb}{https://github.com/jacobgil/pytorch-grad-cam}}.

\subsection{t-SNE Details}

Following the protocol in FGFA \cite{fgfa}, we categorise the ImageNet VID Val set into three groups: fast-speed, medium-speed, and slow-speed subsets.
The definition is based on the Motion Intersection over Union (mIoU) metric which measures the IoU of the same object in the nearby frames (±10 frames). 
The specific thresholds are mIoU $> 0.9$ (slow), mIoU $\in [0.7, 0.9]$ (medium), and mIoU $< 0.7$ (fast).
Slow-speed subset usually has higher quality than the medium-speed and the fast-speed subsets. Therefore, STPN improves the FasterRCNN detector more significantly in the medium-speed and fast-speed subsets.

t-SNE requires a classification label for each sample feature. So we need to convert the feature maps within bounding boxes into sample features. Given a bounding box, we use the RoIPooling \cite{fasterrcnn} operation to generate a sample feature (proposal). Specifically, we use the ground-truth bounding boxes of the ImageNet VID Val set to generate sample features. The labels of each sample feature are set as the label of the corresponding ground-truth bounding box. In this way, we can compare the quality of feature maps obtained by FasterRCNN with and without STPN.

\bibliographystyle{ieee_fullname}
\bibliography{egbib}

\end{document}